# Modelling local and global phenomena with sparse Gaussian processes

**Jarno Vanhatalo**
Department of Biomedical Engineering
and Computational Science
Helsinki University of Technology
02015 TKK, Finland

**Aki Vehtari**
Department of Biomedical Engineering
and Computational Science
Helsinki University of Technology
02015 TKK, Finland

## Abstract

Much recent work has concerned sparse approximations to speed up the Gaussian process regression from the unfavorable $O(n^3)$ scaling in computational time to $O(nm^2)$. Thus far, work has concentrated on models with one covariance function. However, in many practical situations additive models with multiple covariance functions may perform better, since the data may contain both long and short length-scale phenomena. The long length-scales can be captured with global sparse approximations, such as fully independent conditional (FIC), and the short length-scales can be modeled naturally by covariance functions with compact support (CS). CS covariance functions lead to naturally sparse covariance matrices, which are computationally cheaper to handle than full covariance matrices. In this paper, we propose a new sparse Gaussian process model with two additive components: FIC for the long length-scales and CS covariance function for the short length-scales. We give theoretical and experimental results and show that under certain conditions the proposed model has the same computational complexity as FIC. We also compare the model performance of the proposed model to additive models approximated by fully and partially independent conditional (PIC). We use real data sets and show that our model outperforms FIC and PIC approximations for data sets with two additive phenomena.

## 1 Introduction

Gaussian processes (GP) are powerful tools for Bayesian nonlinear and nonparametric regression. They can be used as prior for underlying latent function, on which the observations are conditioned (Rasmussen and Williams, 2006). The main limitation with GP models is their unfavorable $O(n^3)$ scaling in training time and $O(n^2)$ in memory, where $n$ is the size of the training set. In recent years, much research has concerned sparse approximations to speed up the computations down to $O(nm^2)$ and reduce the memory requirements to $O(nm)$, with $m \ll n$ (e.g. Snelson and Ghahramani, 2006, 2007; Lawrence, 2003; Seeger et al., 2003; Williams and Seeger, 2001; Quiñonero-Candela and Rasmussen, 2005). Another approach to speed up training and save memory is to use GPs with compactly supported (CS) covariance functions. These are special kinds of functions that construct naturally sparse covariance matrices (e.g. Wendland, 2005; Rasmussen and Williams, 2006; Storkey, 1999).

In this paper, we will treat both the sparse GP approximations and GPs with naturally sparse covariance function. These two concepts are rather different, but share the property that they are computationally more efficient than full GPs. The sparse approximations considered here are fully and partially independent conditional (FIC/PIC) (Snelson and Ghahramani, 2006, 2007). The naturally sparse GP is discussed for a piecewise polynomial covariance function. FIC provides a global approximation that performs well with rather long length-scales. PIC combines both local and global type of approximations and is able to model also short length-scale phenomena. The problem with PIC is, however, that it introduces discontinuities in the correlation structure. The CS covariance functions, on the other hand, provide a natural way to model the local phenomena without anomalies in the correlation structure.

In many practical problems, it is plausible that the underlying function combines both long and short length-scale phenomena. In this case, we can construct an additive GP model with two covariance functions. To our best knowledge, the sparse GP literature thus far has concentrated on modelling either long or short length-scale phenomena. The purpose of this work is to add these two concepts together to construct a sparse GP model that is able to capture both local and global properties at the same time. Therefore, we propose to add up the covariance function induced by FIC approximation with a CS covariance function so that the long length-scale phenomena are captured by FIC and the local variations by CS function. We will apply the model for real data sets and compare the results to additive full GP models approximated by FIC and PIC. We will show that our method outperforms the approximations in performance and is computationally of same complexity under certain conditions.

## 2 Gaussian process regression

We will consider a regression problem, where we have scalar observations $\mathbf{y} = \{y_i\}_{i=1}^n$ at input locations $\mathbf{X} = \{\mathbf{x}_i\}_{i=1}^n$ and the observations are assumed to satisfy

$$y_i = g(\mathbf{x}_i) + \epsilon, \text{ where } \epsilon \sim \mathcal{N}(0, \sigma^2). \quad (1)$$

Each of the input vectors $\mathbf{x}_i$ is of dimension $D$, which is assumed to be rather low (say $D \leq 3$), and $g$ is a nonlinear latent function, which we are interested in. The latent function is given a Gaussian process prior, which implies that any finite subset of latent variables, $\mathbf{g} = \{g(\mathbf{x}_i)\}_{i=1}^n$, has a multivariate Gaussian distribution. In particular, at the observed input locations $\mathbf{X}$ the latent variables have distribution $p(\mathbf{g}|\mathbf{X}) = \mathcal{N}(\mathbf{g}|\mu, \mathbf{K}_{n,n})$, where $\mathbf{K}_{n,n}$ is the covariance matrix and $\mu$ the mean function. In this paper, without loss of generality, we will use a zero-mean Gaussian processes. The covariance matrix is constructed from a covariance function $k$, which represents the prior assumptions of the smoothness of the latent function. Each element in the covariance matrix is evaluated as $[\mathbf{K}_{n,n}]_{ij} = k(\mathbf{x}_i, \mathbf{x}_j, \theta)$ and a widely used covariance function is the stationary squared exponential

$$k_{\text{se}}(\mathbf{x}_i, \mathbf{x}_j) = \sigma_{\text{se}}^2 \exp\left(-\sum_{d=1}^{D}(x_{i,d} - x_{j,d})^2/l_d^2\right), \quad (2)$$

where $\sigma_{\text{se}}^2$ is the scaling parameter and $l_d$ is the length-scale that governs how fast the correlation drops among direction $d$. From this on we denote the hyperparameters by $\theta = \{\sigma_{\text{se}}^2, l_1, ..., l_D, \sigma^2\}$

Since the noise model, or the likelihood, is Gaussian $p(\mathbf{y}|\mathbf{g}, \mathbf{X}, \theta) = \mathcal{N}(\mathbf{y}|\mathbf{g}, \sigma^2 \mathbf{I})$, we are able to marginalise over the latent variables to obtain the marginal likelihood

$$p(\mathbf{y}|\mathbf{X}, \theta) = \mathcal{N}(\mathbf{y}|\mathbf{0}, \mathbf{K}_{n,n} + \sigma^2 \mathbf{I}). \quad (3)$$

After placing a prior for the hyperparameters $p(\theta)$, we can find the maximum a posteriori (MAP) estimate $\hat{\theta}$ for the hyperparameters by maximising the log posterior cost function,

$$\hat{\theta} = \arg\max_{\theta}\left[\log p(\theta) - \frac{1}{2}\log|\mathbf{K}_{y,y}| - \frac{1}{2}\mathbf{y}^{\text{T}}\mathbf{K}_{y,y}^{-1}\mathbf{y}\right], \quad (4)$$

where $\mathbf{K}_{y,y} = \mathbf{K}_{n,n} + \sigma^2\mathbf{I}$. Optimising the hyperparameters can also be considered as a model selection, since we are fixing our GP model by setting the hyperparameters in their MAP estimate.

Conditioning on the data and the hyperparameter values, the posterior predictive distribution of the latent variables $g(\mathbf{X}_*)$ at new input locations $\mathbf{X}_*$ is Gaussian

$$p(\mathbf{g}_*|\mathbf{X}_*, \mathcal{D}, \hat{\theta}) = \mathcal{N}(\mathbf{g}_*|\mu_*, \mathbf{\Sigma}_{*,*}), \quad (5)$$

where $\mathcal{D} = \{\mathbf{y}, \mathbf{X}\}$ and

$$\mu_* = \mathbf{K}_{*,n}\mathbf{K}_{y,y}^{-1}\mathbf{y}$$
$$\mathbf{\Sigma}_{*,*} = \mathbf{K}_{*,*} - \mathbf{K}_{*,n}\mathbf{K}_{y,y}^{-1}\mathbf{K}_{n,*}. \quad (6)$$

In many practical situations one length-scale per input dimension is too restrictive. Consider, for example, time series data, which evolves rather smoothly between different years, but at the same time has faster, monthly, variations (see figure 1). In this case, a more reasonable model is

$$y_i = g(\mathbf{x}_i) + \epsilon = f(\mathbf{x}_i) + h(\mathbf{x}_i) + \epsilon, \quad (7)$$

where the latent function $g$ is replaced by a sum of two functions, of which the other is slowly and the other fast varying. We can now place a Gaussian process prior for both of the functions $f$ and $h$ and give them different covariance functions that reflect our beliefs about their smoothness properties. The sum of two Gaussian variables is Gaussian and the prior for the additive model is $p(\mathbf{g}|\mathbf{X}) = \mathcal{N}(\mathbf{g}|\mathbf{0}, \mathbf{K}_{n,n}^{(h)} + \mathbf{K}_{n,n}^{(f)})$. The marginal likelihood and posterior predictive distribution are as before with $\mathbf{K}_{n,n} = \mathbf{K}_{n,n}^{(h)} + \mathbf{K}_{n,n}^{(f)}$. However, if we are interested on only, say, phenomenon $f$, we can consider the $h$ part of the latent function as correlated noise and evaluate the predictive distribution $p(f|\mathbf{X}_*, \mathcal{D}, \hat{\theta}) = N(\mathbf{K}_{*,n}^{(f)}\mathbf{K}_{y,y}^{-1}\mathbf{y}, \mathbf{K}_{*,*}^{(f)} - \mathbf{K}_{*,n}^{(f)}\mathbf{K}_{y,y}^{-1}\mathbf{K}_{n,*}^{(f)})$.

The training of the hyperparameters is conducted via gradient based optimisation. The computationally most demanding part is the evaluation of the gradient of the log marginal likelihood

$$\frac{\partial}{\partial \theta}\log p(\mathbf{y}|\mathbf{X}, \theta) = \frac{1}{2}\mathbf{y}^{\text{T}}\mathbf{K}_{y,y}^{-1}\frac{\partial \mathbf{K}_{y,y}}{\partial \theta}\mathbf{K}_{y,y}^{-1}\mathbf{y}$$
$$- \frac{1}{2}\text{tr}\left(\mathbf{K}_{y,y}^{-1}\frac{\partial \mathbf{K}_{y,y}}{\partial \theta}\right). \quad (8)$$

The matrix inversion and the determinant in marginal likelihood and its gradient scale as $O(n^3)$ in time. Together with $O(n^2)$ memory requirements this prevents the direct implementation of GP for large problems.

## 3 FIC and PIC sparse approximations

In this section, we consider the fully and partially independent conditional sparse approximations for GP. We will give a short review of the methods, but readers interested in detailed derivation of the approximations should refer to the original papers by Snelson and Ghahramani (2006, 2007), or for a more general perspective to Quiñonero-Candela and Rasmussen (2005).

The approximations are based on introducing an additional set of latent variables $\mathbf{u} = \{u_i\}_{i=1}^m$, called *inducing variables*, that correspond to a set of input locations $\mathbf{X}_u$, *inducing inputs*. The prior on latent function is approximated by

$$p(\mathbf{g}|\mathbf{X}) \approx q(\mathbf{g}|\mathbf{X}, \mathbf{X}_u) = \int q(\mathbf{g}|\mathbf{X}, \mathbf{X}_u, \mathbf{u}) p(\mathbf{u}|\mathbf{X}_u) d\mathbf{u}, \quad (9)$$

where $\mathbf{g}$ is conditionally dependent on $\mathbf{u}$ through the inducing conditional $q(\mathbf{g}|\mathbf{X}, \mathbf{X}_u, \mathbf{u})$. In the FIC framework, the latent variables are conditionally independent given $\mathbf{u}$. In this case, the inducing conditional factorises into $q(\mathbf{g}|\mathbf{X}, \mathbf{X}_u, \mathbf{u}) = \prod_{i=1}^n q_i(g_i|\mathbf{X}, \mathbf{X}_u, \mathbf{u})$. In contrast, in PIC the latent variables are set into blocks that are conditionally independent, but within a block they are fully dependent. The approximate conditionals of FIC and PIC can be summarised as

$$q(\mathbf{g}|\mathbf{X}_u, \mathbf{u}) = \mathcal{N}(\mathbf{K}_{n,u} \mathbf{K}_{u,u}^{-1} \mathbf{u}, \mathrm{mask}(\mathbf{K}_{n,n} - \mathbf{Q}_{n,n}, \mathbf{M})),$$

where $\mathbf{Q}_{a,b} = \mathbf{K}_{a,u} \mathbf{K}_{u,u}^{-1} \mathbf{K}_{u,b}$ and $\mathbf{K}_{n,u}$ is the covariance between latent variables and the inducing variables. The function $\mathrm{mask}(\mathbf{K}, \mathbf{M})$, with matrix $\mathbf{M}$ of ones and zeros, returns a matrix $\mathbf{\Lambda}$ of size $\mathbf{M}$ and elements $\mathbf{\Lambda}_{ij} = \mathbf{K}_{ij}$ if $\mathbf{M}_{ij} = 1$ and zero otherwise. An approximation with $\mathbf{M} = \mathbf{I}$ corresponds to FIC and an approximation, where $\mathbf{M}$ is block diagonal corresponds to PIC.

By placing a zero-mean Gaussian prior on the inducing variables, $\mathbf{u} \sim \mathcal{N}(\mathbf{0}, \mathbf{K}_{u,u})$, and integrating over them we get the approximate prior over latent variables

$$q(\mathbf{g}|\mathbf{X}, \mathbf{X}_u) = \mathcal{N}(\mathbf{0}, \mathbf{Q}_{n,n} + \mathbf{\Lambda}). \quad (10)$$

Here the inducing inputs can be considered as an additional set of hyperparameters. The marginal likelihood is $p(\mathbf{y}|\mathbf{X}, \mathbf{X}_u, \theta) = \mathcal{N}(\mathbf{0}, \mathbf{Q}_{n,n} + \mathbf{\Lambda} + \sigma^2 \mathbf{I})$, where the covariance is now a sum of the rank $m$ matrix $\mathbf{Q}_{n,n}$ and (block)diagonal matrix $\mathbf{\Lambda} + \sigma^2 \mathbf{I}$. This leads to computational savings in training, since the crucial computations only require $O(nm^2)$. Other major advantage is the save in memory requirements, since the sparse approximations need only $O(nm)$ memory compared to $O(n^2)$ in full GP.

In the literature on FIC and PIC, the approximations are used for the one covariance function regression problem (1). The results show that the approximations are efficient for a wide variety of such problems (Snelson, 2007). However, the two covariance function regression problem (7) is somewhat trickier as will be seen in the experiments. With FIC all the correlations are induced through the inducing inputs. Thus, modelling very short length-scale phenomena becomes unpractical, since one would need infeasible many inducing inputs to capture fast variations. PIC, on the other hand, can capture short length-scales within a block, but its shortcoming is that between the blocks latent variables are independent. This leads to discontinuities in correlation structure and reduces the predictive performance on the block boundaries.

## 4 Gaussian process with compact support covariance function

In this section, we will consider Gaussian processes with naturally sparse covariance matrix. We will first give short introduction on compact support covariance function, after which we consider their implementation issues within GP framework.

### 4.1 Compact support

With a compactly supported covariance function we mean a function that gives zero correlation between data points whose distance exceeds a certain threshold. Such functions form naturally sparse covariance matrices, that give savings in computational time and memory requirements compared to full GPs. The challenge in constructing CS covariance functions is to guarantee their positive definiteness. A full support covariance function can not be cut arbitrarily to obtain a compact support, since the resulting function would not, in general, be positive definite. One option is to use a family of piecewise polynomial functions as, for example,

$$k_{\mathrm{pp}} = \frac{\sigma_{\mathrm{pp}}^2}{3}(1-r)_+^{j+2}\left((j^2+4j+3)r^2 + (3j+6)r + 3\right), \quad (11)$$

where $j = \lfloor D/2 \rfloor + 3$ and $r^2 = \sum_{d=1}^D (x_{i,d} - x_{j,d})^2/l_d^2$ (see e.g. Wendland, 2005; Rasmussen and Williams, 2006). The function $k_{\mathrm{pp}}$ is positive definite up to the input dimensions $D$. There are many other CS covariance functions and, for example, Gneiting (2002) proposed a method that constructs a CS covariance function that preserves the smoothness properties of powered exponential and Matern class of functions.

In this work, we concentrate on, how GP analysis can efficiently be done with CS functions and leave the comparison of different functions for future. The CS covariance functions in relation to GPs have been studied already by, for example, Storkey (1999). He considered covariance matrices of Toeplitz form, which are fast to handle due their banded structure. However, constructing Toeplitz covariance matrices is not possible in two or higher dimensions without approximations. Here we will present an approach that works for any kind of sparse covariance matrix.

The computationally intensive parts in the cost function (4) and its derivatives (8) are the determinant and multiplication of matrix or vector by the inverse covariance matrix. Albeit the covariance matrix is sparse its inverse is usually full and requires time $O(n^3)$ to compute. Fortunately, we are able to do all these calculations without forming the full inverse. The key role is played by the Cholesky factorisation of a covariance matrix.

### 4.2 The computations

The Cholesky factorization $\mathbf{L}$ of symmetric positive definite matrix $\mathbf{A}$ is a lower triangular matrix such that $\mathbf{A} = \mathbf{L}\mathbf{L}^T$. For full matrix, it can be found in time $O(n^3)$. For sparse matrices, however, this is faster since the sparsity is preserved in the Cholesky factorisation. The factorisation time depends on the sparsity structure of the matrix $\mathbf{A}$. It can be reduced by permuting the columns and rows of matrix so that the number of non-zero elements in $\mathbf{L}$ are minimised (for example, using minimum degree ordering or nested dissection (Amestoy et al., 2004; Davis, 2006; Rue and Martino, 2007)). The time needed for permutation is, in general, negligible compared to the time needed for other evaluations with GP. For example, in $N \times N$ lattices and with small length-scale in CS function the typical cost of factorising CS covariance matrix after permutation is $O(n^{3/2})$ for $n = N^2$ and for $N \times N \times N$ lattices $O(n^2)$, for $n = N^3$. After finding the Cholesky decomposition of matrix $\mathbf{A}$, we can efficiently evaluate its log determinant and $\mathbf{y}^T \mathbf{A} \mathbf{y}$ ($O(n)$ and $O(n \log n)$ operations with 2D lattice data).

The term that needs most concern is $\mathrm{tr}(\mathbf{K}_{y,y}^{-1} \partial \mathbf{K}_{y,y}/\partial \theta)$ in the gradient of the marginal likelihood (8). The matrix $\mathbf{B} = \partial \mathbf{K}_{y,y}/\partial \theta$ has the same sparsity structure as $\mathbf{K}_{y,y}$. Thus, if we denote $\mathbf{Z} = \mathbf{K}_{y,y}^{-1}$ we can write $\mathrm{tr}(\mathbf{Z}\mathbf{B}) = \mathrm{tr}(\mathbf{Z}_{\mathrm{sp}}\mathbf{B})$, where $\mathbf{Z}_{\mathrm{sp}}$ is a sparse representation of $\mathbf{Z}$, which has non-zero elements only where $\mathbf{B}_{ij} \neq 0$. We can obtain such a matrix by using an algorithm introduced by Takahashi et al. (1973).

The algorithm is derived as follows. First, determine the Cholesky decomposition of $\mathbf{K}_{y,y}$ and write

$$\mathbf{L}^T \mathbf{Z} = \mathbf{L}^{-1}. \qquad (12)$$

Next, take the diagonal of the Cholesky triangle, $\mathbf{D} = \mathrm{mask}(\mathbf{L}, \mathbf{I})$, write the equation (12) as

$$\mathbf{D}\mathbf{Z} + (\mathbf{L}^T - \mathbf{D})\mathbf{Z} = \mathbf{L}^{-1}, \qquad (13)$$

subtract the second term on the left hand side and multiply by $\mathbf{D}^{-1}$ from left

$$\mathbf{Z} = \mathbf{D}^{-1}\mathbf{L}^{-1} - \mathbf{D}^{-1}(\mathbf{L}^T - \mathbf{D})\mathbf{Z}. \qquad (14)$$

Now, since the inverse is symmetric, we can give a recursive formula for the elements of the inverse as

$$\mathbf{Z}_{ij} = \frac{\delta_{ij}}{\mathbf{D}_{ii}^2} - \frac{1}{\mathbf{D}_{ii}} \sum_{k=i+1}^{n} \mathbf{L}_{ki} \mathbf{Z}_{kj}, \; j \geq i, i = n, ..., 1. \quad (15)$$

We can find the upper triangle of the inverse by looping $i$ from $n$ to 1 and $j$ from $n$ to $i$. The lower triangular of $\mathbf{Z}$ can then be filled according to symmetry. To find the sparse inverse we evaluate only a small fraction of the elements.

Let us denote by $\mathbf{C}$ an adjacency matrix, which has $\mathbf{C}_{ij} = 1$ if $\mathbf{L}_{ij} \neq 0$ or $\mathbf{L}_{ij}^T \neq 0$ and zero otherwise. Here we consider $\mathbf{L}_{ij}$ non-zero even if its numerical value was zero, but it is symbolically non-zero (that is, it has to be evaluated, when solving for sparse Cholesky factorisation (Davis, 2006)). To find the sparse inverse, we need to evaluate only the elements $\mathbf{Z}_{ij}$ such that $\mathbf{C}_{ij} \neq 0$. To justify this, consider the recursive formula (15). The algorithm needs at least all the elements $\{\mathbf{Z}_{ij} : \mathbf{C}_{ij} \neq 0\}$, since these are the elements that are needed for the diagonal of $\mathbf{Z}$. From the Cholesky factorisation it follows that $\mathbf{C}_{ij}$ is non-zero, if for any $k < i, j$ the elements $\mathbf{C}_{ik}$ and $\mathbf{C}_{kj}$ are non-zero:

$$k < i, j \; \mathbf{C}_{ik} \neq 0 \; \mathbf{C}_{kj} \neq 0 \Rightarrow \; \mathbf{C}_{ij} \neq 0.$$

This property ensures that using the equation (15) we are able to find all the elements $\mathbf{Z}_{ij}$, for which $\mathbf{C}_{ij} \neq 0$, by looping $i$ from $n$ to 1 and at each $i$ evaluating the elements $\mathbf{Z}_{ij}, j \in c_i = \{j \geq i, C_{ij} \neq 0\}$. The algorithm is discussed in detail by Niessner and Reichert (1983).

### 4.3 Computational complexity

The computational complexity and memory requirements of CS+FIC depend on the number of (symbolically) nonzero elements in the Cholesky factorisation of $\mathbf{K}_{y,y}$. This is always equal or more than the number of non-zeros in the covariance matrix and with 2D and 3D lattice data it is $O(n \log n)$ and $O(n^{4/3})$ (Davis, 2006). If $\gamma_k$ denotes the number of non-zero elements outside the diagonal of column $k$ of $\mathbf{L}$, then the computational cost for finding the sparse inverse is $O(\sum_{k=1}^{n} \gamma_k(\gamma_k+1))$. To get an intuition of the computation time we can consider banded and full covariance matrices as limiting cases. Finding the sparse inverse of banded covariance matrix needs time $O(n(b/2)^2)$, where $b$ denotes the band width, and finding the full inverse is $O(n^3/3)$ operation. The other sparse inverses place between these. For example, the average

$\gamma_k$ of the two and three dimensional lattice data are $O(\log n)$ and $O(n^{1/3})$. Thus the recursion takes time $O(n \log(n)^2)$ and $O(n^{5/3})$, respectively. For randomly distributed data points this is an approximation.

## 5 Additive model with CS covariance function and FIC

Here, we propose a new sparse GP model that combines the ideas behind sparse approximations and CS covariance functions.

### 5.1 Additive sparse GP model

The covariance matrix in the FIC approximate prior (10) can be interpreted as a realisation of special kind of covariance function $k_{\text{FIC}} = f(k(\mathbf{x}_i, \mathbf{x}_j, \theta), \mathbf{X}_u)$. This is a function of the original covariance function, its hyperparameters and the inducing inputs. By adding up the FIC and CS covariance functions we are able to construct a sparse GP model for the two component regression problem (7) with prior

$$p(\mathbf{g}|\mathbf{X}, \mathbf{X}_u, \theta) = \mathcal{N}(\mathbf{0}, \mathbf{Q}_{n,n} + \mathbf{\Lambda} + \mathbf{K}_{n,n}^{(\text{CS})}). \quad (16)$$

We will refer to this later as CS+FIC model. Here, the matrix $\mathbf{Q}_{n,n}$ is of rank $m$ and the matrix $\mathbf{\Lambda} + \mathbf{K}_{n,n}^{(\text{CS})}$ is sparse with the same sparsity structure as in $\mathbf{K}_{n,n}^{(\text{CS})}$. Now, we can conduct the training in similar manner to FIC and PIC by using the Woodbury-Sherman-Morrison lemma

$$(\mathbf{Q}_{n,n} + \hat{\mathbf{\Lambda}})^{-1} = \hat{\mathbf{\Lambda}}^{-1} - \mathbf{V}\mathbf{V}^{\text{T}}, \quad (17)$$

where

$$\mathbf{V} = \hat{\mathbf{\Lambda}}^{-1} \mathbf{K}_{g,u} \text{chol}[(\mathbf{K}_{u,u} + \mathbf{K}_{u,g} \hat{\mathbf{\Lambda}}^{-1} \mathbf{K}_{g,u})^{-1}].$$

In case of PIC and FIC $\hat{\mathbf{\Lambda}} = \mathbf{\Lambda} + \sigma^2 \mathbf{I}$ and in CS+FIC $\hat{\mathbf{\Lambda}} = \mathbf{K}_{n,n}^{(\text{CS})} + \mathbf{\Lambda} + \sigma^2 \mathbf{I}$. In FIC and PIC, most of the time is spent in the matrix multiplications in $\mathbf{V}$. The inverse of $\mathbf{\Lambda} + \sigma^2 \mathbf{I}$ can be evaluated in $O(n)$, with FIC, and in $O(nb^2)$, with PIC, where $b$ is the blocksize. After this, the rest of the computations required in the log marginal likelihood and its gradients involve sums and products of $\hat{\mathbf{\Lambda}}^{-1}$ and $\mathbf{V}$ with (block)diagonal matrices and matrices of size at most $n \times m$ (for technical details see, e.g., Snelson (2007))[1].

### 5.2 Computational issues

The matrix multiplications between $n \times m$ matrices are present also in CS+FIC model, and, thus, its computational time is at least the same as for FIC and PIC. The interesting question, however, is what is the upper bound for the computational complexity of CS+FIC. The terms that need consideration are those involving $\hat{\mathbf{\Lambda}}$. The multiplications $\hat{\mathbf{\Lambda}}^{-1} \mathbf{P}$, where $\mathbf{P}$ is an $n \times m$ matrix (for example, $\mathbf{K}_{g,u}$ in $\mathbf{V}$), are computed by solving $m$ linear equations, and the terms $\text{tr}(\hat{\mathbf{\Lambda}}^{-1} \partial \mathbf{H}/\partial \theta)$, where $\mathbf{H}$ has the same sparsity structure as $\hat{\mathbf{\Lambda}}$, are solved using the sparse inverse algorithm described earlier. If we assume uniformly distributed 2D lattice data and small length-scale for CS function, we need $O(n^{3/2})$ time for the Cholesky factorisation of $\hat{\mathbf{\Lambda}}$, $O(nm \log(n))$ for solving the $m$ linear equations and $O(n \log(n)^2)$ for the sparse inverse. With three dimensional lattice, the computations are governed by the $O(n^2)$ scaling of the Cholesky factorisation. We can now conclude that the computational complexity of the CS+FIC model is: $O(nm^2) \leq O_{\text{CS+FIC}} \leq O(\max(nm^2, n \log(n)^2, n^{3/2}))$ for 1D and 2D lattice. For the 3D lattice the performance is $O(nm^2) \leq O_{\text{CS+FIC}} \leq O(\max(nm^2, n^2))$. The memory requirements are $O(n \log(n))$ and $O(n^{4/3})$ respectively. Thus, if $m \geq \log(n)$ and $m \geq \sqrt{n}$ the training of CS+FIC model scales up to a constant as the training of FIC (For example 100 inducing inputs and 10 000 data points).

## 6 Experiments

In this section, we present results on applying the models discussed above to several data sets.

### 6.1 Maunaloa $CO_2$ data

The data contain the atmospheric $CO_2$ concentrations (ppmv) collected at Mauna Loa Observatory, Hawaii, every month from 1958 until 2004[2] (see figure 1(a)). The data consist of 557 points (seven measurements were missing) and were previously analysed, for example, by Rasmussen and Williams (2006). They presented a sophisticated model that was able to capture the periodicity, long term trend and changes in the periodicity. Here, the aim is to test sparse additive GP models. The models compared are: full GP, FIC and PIC with $k_{\text{se}} + k_{\text{pp}}$ covariance function, and CS+FIC model, where the CS part is given by $k_{\text{pp}}$ and FIC part uses $k_{\text{se}}$. We placed 24 inducing inputs in a regular grid over the input space. Since FIC failed with 24 inducing inputs, we ran additional experiments with 141 inducing inputs with FIC. The blocks for PIC were set regularly so that the number of data points in a block was approximately same as the number of inducing inputs. The length-scales were given half Students'-$t$ prior with 3 degrees of freedom and variance 4 and

---

[1] The Matlab implementation of the models used in this work will be published in www.lce.hut.fi/research/mm/gp/

[2] http://cdiac.esd.ornl.gov/ftp/trends/co2/maunaloa.co2

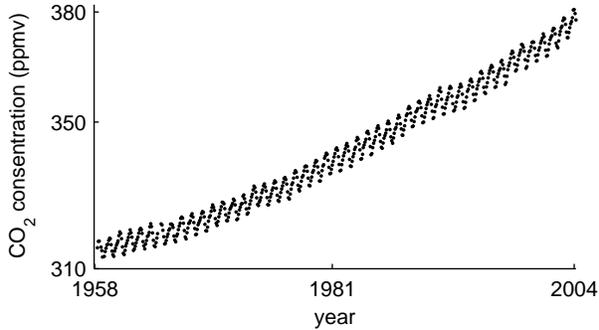

(a) The Maunaloa $CO_2$ data.

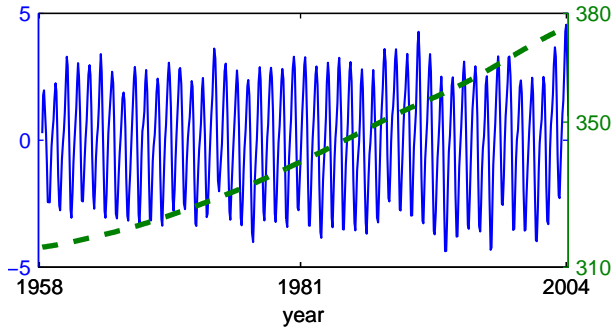

(b) The posterior predictive mean of the monthly and overall trend with CS+FIC. On the left the scale for the short term trend and on the right for the long term trend.

Figure 1: The Maunaloa $CO_2$ data and predictions with CS+FIC model.

Table 1: The model performances in the Maunaloa $CO_2$ data obtained with 10-fold cross validation.

| Model | RMSE | MLPD |
|---|---|---|
| CS+FIC (m=24) | 0.317 | -0.251 |
| full GP | 0.316 | -0.250 |
| FIC (m=24) | 2.151 | -2.189 |
| FIC (m=141) | 0.83 | -1.265 |
| PIC (m=24) | 0.401 | -0.318 |

the magnitudes half Students'-$t$ with 0.3 degrees of freedom and variance 4.

The predictive mean of the CS+FIC model for the short and long term behaviour are shown in the figure 1(b) and the model performances are presented in the table 1. We conducted 10-fold cross-validation and evaluated the root mean squared error (RMSE) and the mean log predictive density (MLPD) for each of the models. The CS+FIC was as good as the full GP in both of the performance tests. PIC was little worse than CS+FIC, which can be explained by the discontinuities on the block boundaries. FIC with 24 inducing inputs did orders of magnitude worse than all the other models, because it was able to capture only the long trend phenomenon and considered the monthly changes as noise. With 141 inducing inputs FIC did better, since it was able to model also the monthly changes. However, it was still worse than the other models.

## 6.2 The US annual precipitation data

The US precipitation data consist of monthly precipitation measures recorded across the whole country from 1895 to 1997[3]. The data consist of the spatial co-ordinates and elevation of the stations, and the precipitation in millimeters per month. In total, there are 11918 stations, but a high fraction of the measurements are missing. For the analysis in this paper, we collected the stations that recorded all the measurements for the year 1995. This leads to total 5776 stations (the station locations are shown in figure 2(a)). The data was previously used by Paciorek and Schervish (2006), who studied the subregion of Colorado with non-stationary covariance function. However, they recorded equally good results for stationary covariance functions, which suggests that an additive stationary model could work reasonably well also for the whole country.

We will study the data first with a model that uses only the spatial co-ordinates as inputs for GP. After this we study the same data with all three explanatory variables. Since the number of data points is infeasible high for full GP, we tested only FIC, PIC and CS+FIC with the same covariance functions as in the previous experiment. The inducing inputs were treated in two different ways. First, they were initialised by picking 90 data points (for FIC we used also 225 inducing inputs) randomly from the data and optimized. This gave equally good results compared to the model, where they were placed in regular lattice over the country. Thus, we will report only the results from the latter scheme. The blocks for the PIC model were placed regularly over the country so that the number of data points inside each block was approximately the same as the number of inducing inputs. For the training, the input co-ordinates were scaled between 1-120 and the targets between -9-35. The priors for hyperparameters were the same as in the previous experiment.

The model performances were evaluated via 10-fold cross validation. Results are shown in the tables 2 and 3. The posterior predictive mean of the precipitation levels across the country is shown in figure 2(b). Again in this data set, CS+FIC outperforms the FIC

---

[3] http://www.image.ucar.edu/GSP/Data/US.monthly.met/

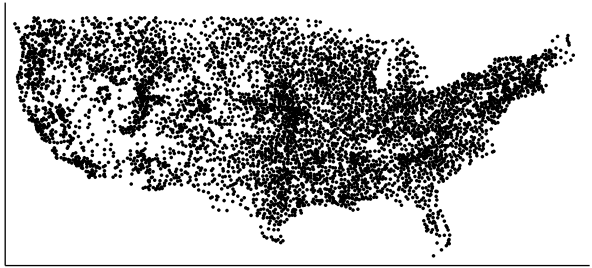

(a) The station locations.

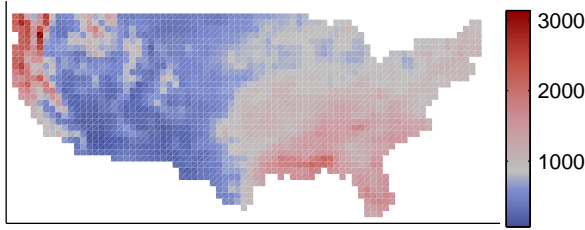

(b) The annual precipitation in the US in 1995.

Figure 2: The US annual precipitation data. The upper figure shows the data points and the lower figure shows the CS+FIC posterior mean of the annual precipitation in the US in 1995.

Table 2: The model performances in the US annual precipitation data with *spatial covariates* obtained with 10-fold cross validation.

| Model | RMSE | MLPD |
|---|---|---|
| CS+FIC (m=90) | 208.3 | -2.164 |
| FIC (m=90) | 278.1 | -2.444 |
| FIC (m=225) | 263.0 | -2.394 |
| PIC (m=90) | 210.3 | -2.175 |

and PIC models, but not as clearly as in Maunaloa dataset. The precipitation level varies rather smoothly across most of the US and, thus, FIC and PIC are able to capture the behaviour in large areas in the east and middle of the country. The quick changes in the precipitation level take place in the mountainous regions in the west. These changes can be explained by the elevation information, since the models with elevation information have considerably better predictive performance than the models without.

### 6.3 Computational performance

In the Maunaloa data set, the training time of CS+FIC was the same as the training time of FIC/PIC. In this case the data was collected with constant rate, which resulted in banded (Toeplitz form) $\hat{\mathbf{\Lambda}}$, whose Cholesky factorisation is very sparse. In the precipitation data sets, CS+FIC was little slower than FIC/PIC, but still the training time was only few minutes with standard office PC. The matrix $\hat{\mathbf{\Lambda}}$ had less than 3% non-zero elements in all the data sets and the memory requirements of CS+FIC were the same as in FIC/PIC.

Table 3: The model performances in the US annual precipitation data with *spatial and elevation* covariates obtained with 10-fold cross validation.

| Model | RMSE | MLPD |
|---|---|---|
| CS+FIC (m=90) | 165.2 | -1.788 |
| FIC (m=90) | 249.1 | -2.321 |
| FIC (m=225) | 216.9 | -2.189 |
| PIC (m=90) | 168.7 | -1.898 |

The scaling of the training time of CS+FIC was tested by keeping the amount of inducing inputs and data points constant at turn and altering the other. With Maunaloa data set CS+FIC scaled exactly as FIC and PIC. With the precipitation data sets the training time of CS+FIC with fixed number of inducing inputs increased slightly faster than theoretically. The reasons for this are non uniformly collected data and the extra overhead in the sparse inverse algorithm for keeping track of the non-zero elements in $\mathbf{Z}_{\text{sp}}$. The sparse inverse algorithm is currently implemented with Matlab, which results in rather high overhead in the for-loops. Also, the short length-scale phenomena might not have been present in the subsampled data sets. This increases the sparsity of $\hat{\mathbf{\Lambda}}$ and speeds up the CS+FIC model. When comparing the computation times, one has to remember that in contrast to FIC/PIC the training time of CS+FIC depends on the characteristics of the data. Thus, direct comparison of the training times is hard.

Albeit we have considered only low dimensional problems, nothing prevents us to ably CS+FIC for high dimensional problems as well. In general, the density of the sparse covariance matrix tends to increase as the input dimension gets higher. If the data are evenly distributed over the space, the number of data points covered by constant length-scale increases together with dimensionality. However, in order to find fast varying phenomena the data has to be densely distributed across all dimensions. High dimensional regression problems often seem to have broad trends, partly because the density of the data is so low that the fast phenomena can not be found. On request of the reviewers we tested our method also in two data sets, with higher dimensionality than three. The data sets were forest fires data[4], with 12 inputs and 517 data points, and add10 data[5], with ten inputs and 5492 data points. The forest fires data did not contain additive phenomena and the model performance was equally good with all the models. The training times were the same as in Maunaloa data. The add10 data consists of additive components that are functions of different inputs. In this data the CS+FIC worked slightly better and was as fast as in the US precipitation data sets. Due to space constraints the exact results were left out.

---
[4] http://archive.ics.uci.edu/ml/datasets/Forest+Fires

## 7 Conclusions

In this paper, we introduced an additive sparse Gaussian process that can model both long and short length-scale phenomena in the data. The model consists of FIC sparse approximation, which is used for the global phenomenon, and a compact support covariance function to model the local behavior. The analysis of the proposed model is conducted using sparse matrix routines and sparse inverse algorithm introduced by Takahashi et al. (1973). Under certain conditions the computational time is shown to scale similarly to the training time of FIC and PIC.

The CS+FIC model was compared to FIC and PIC approximations with several data sets. We found that the proposed model gives better overall performance than FIC and PIC, if there are two additive phenomena in the data, and equally good performance in non-additive data sets. Approximating additive GP models with short length-scale phenomena by FIC leads to poor performance, because the approximation is global by its nature. PIC models rather well also short length-scales, but its short coming are the discontinuities in the correlation structure. Our model combines the good global properties of FIC and the good local properties of compact support covariance function.

In practical problems it is often reasonable to use additive models for the purposes of data analysis. The proposed CS+FIC model provides a practical tool for modeling large data sets, which are infeasible for full Gaussian processes. The CS+FIC model is faster and requires less memory than full GP. It is also more accurate than FIC and PIC in additive models.

**Acknowledgments**

Authors thank Marc Deisenroth for excellent comments on the paper. The first author thanks also the Graduate School in Electronics, Telecommunications and Automation (GETA) and Finnish Funding Agency for Technology and Innovation for funding his post graduate studies and this research.

---

[5]http://www.cs.toronto.edu/ delve/data/add10/desc.html